\pdfoutput=1
\documentclass[journal,twocolumn]{IEEEtran}

\usepackage{graphicx}
\usepackage{times}
\usepackage{epsfig}
\usepackage{amsmath}
\usepackage{amssymb}
\usepackage{balance}
\usepackage[dvipsnames]{xcolor}
\newcommand{\bestperformance}{\textcolor{red}}
\newcommand{\secondbestperformance}{\textcolor{blue}}
\newcommand{\revision}{\textcolor{black}}
\newcommand{\etal}{\textit{et al. }}
\begin{document}

\title{A Self-Reasoning Framework for Anomaly Detection Using Video-Level Labels} 

\author{Muhammad Zaigham Zaheer, Arif Mahmood, Hochul Shin and Seung-Ik Lee
\thanks{This work is supported by the ICT R\&D program of MSIP/IITP. [2017-0-00306, Dvlp. of Multimodal Sensor-based Intll. Sys. for Outdoor Surveillance Robots]. We thank Jin-ha Lee \& Marcella Astrid for their dedicated help.}
\thanks{M. Zaigham Zaheer, Hochul Shin and Seung-Ik Lee are with the Electronics and Telecommunications Research Institute(ETRI) and the University of Science and Technology (UST), Daejeon, Korea. mzz@ust.ac.kr, \{creatrix, the\_silee\}@etri.re.kr.}
\thanks{Arif Mahmood is with the Information Technology University, Lahore, Pakistan. rfmahmood@gmail.com.}\vspace{-6mm}
}

\twocolumn[
  \begin{@twocolumnfalse}
        This is a draft version of the article accepted by the IEEE Signal Processing Letters.
        \copyright  2020 IEEE. Personal use of this material is permitted. Permission from IEEE must be obtained for all other uses, in any current or future media, including reprinting/republishing this material for advertising or promotional purposes, creating new collective works, for resale or redistribution to servers or lists, or reuse of any copyrighted component of this work in other works. 
 \maketitle

 \end{@twocolumnfalse}
 ]

 \markboth{\copyright  2020 IEEE. IEEE SIGNAL PROCESSING LETTERS, Vol. X, No. Y, July 2020}
{Shell \MakeLowercase{\textit{et al.}}: Bare Demo of IEEEtran.cls for IEEE Journals}
\begin{abstract}
Anomalous event detection in surveillance videos is a challenging and practical research problem among image and video processing community.
Compared to the frame-level annotations of anomalous events, obtaining video-level annotations is quite fast and cheap though such high-level labels may contain significant noise. More specifically, an anomalous labeled video may actually contain anomaly only in a short duration while the rest of the video frames may be normal.
In the current work, we propose a weakly supervised anomaly detection framework based on deep neural networks which is trained in a self-reasoning fashion using only video-level labels. To carry out the self-reasoning based training, we generate pseudo labels by using binary clustering of spatio-temporal video features which helps in mitigating the noise present in the labels of anomalous videos. Our proposed formulation encourages both the main network and the clustering to complement each other in achieving the goal of more accurate anomaly detection. The proposed framework has been evaluated on publicly available real-world anomaly detection datasets including UCF-crime, ShanghaiTech and UCSD Ped2. The experiments demonstrate superiority of our proposed framework over the current state-of-the-art methods.
\end{abstract} 


\IEEEpeerreviewmaketitle

\section{Introduction}

\IEEEPARstart{A}{nomalous} event detection is a challenging problem in signal, image and video processing areas because of its applications in real-world surveillance systems \cite{mestav2020universal,sultani2018real,xiao2015learning,doshi2020fast}. 
Infrequent occurrences make the anomalous events to usually appear as outliers from the normal behavior \cite{chan2008ucsd,ravanbakhsh2017abnormal,lu2013abnormal,bergmann2020uninformed}. Therefore, anomaly detection has often been carried out using one-class classifiers which learn the frequently occurring events as normal \cite{Nguyen_2019_ICCV,Gong_2019_ICCV,zaheer2020old,sabokrou2018ALOCC}. Anomalies are then detected based on their deviations from the learned representations of the normal class. However, it is not usually feasible to collect a complete set of all possible normal scenarios for training, therefore new occurrences of the normal class may also substantially differ from the learned representations and may be misdetected as anomalous \cite{sultani2018real}.

Another widely used anomaly detection approach is to utilize the weakly supervised learning paradigm to train a binary classifier using both the normal and the anomalous data instances \cite{sultani2018real,zhong2019graph,zhang2019temporal}. In such setting, all events in a normal labeled  video are marked as normal. Whereas in an abnormal labeled video, though some of the events may actually be normal, all events are marked as anomalous resulting in noisy labels. It reduces the efforts required in obtaining detailed manual annotations of the dataset, however the training using such type of labels is quite challenging. 
In the current work, we propose an approach for anomalous event detection using such video-level labels.

\begin{figure*}[t]
\begin{center}
\includegraphics[width=0.95\linewidth]{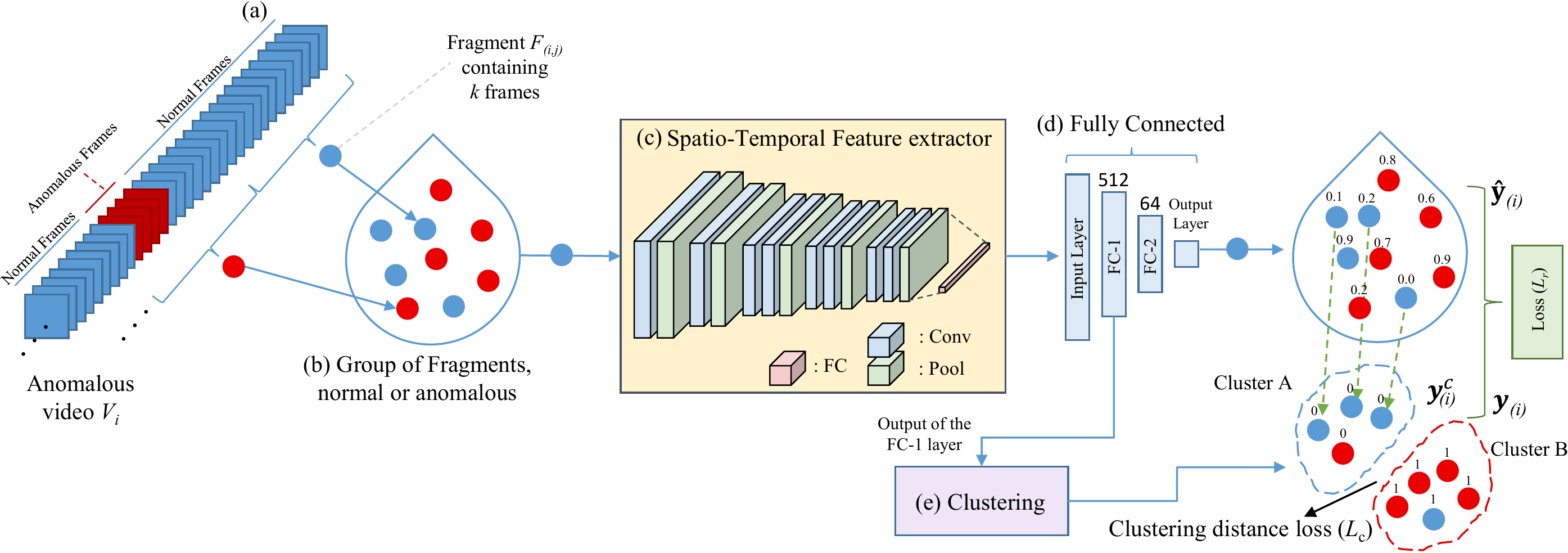}
\end{center}
  \caption{Our proposed architecture for anomaly detection in self-reasoning weakly supervised setting. The labels are provided only at video-level. The video frames (a) are converted into a group of fragments (b). Feature extraction is performed on each fragment (c) and the features are input to the FC network (d). Intermediate representations of a whole video inferred from the FC-1 layer are used to create clusters (e). For an anomalous labeled video, the pseudo labels $y^p$ are generated with the help of clusters.}
\label{fig:architecture}
\vspace{-2mm}
\end{figure*}

Recently, anomaly detection problem in weakly labeled videos has been formulated as Multiple Instance Learning (MIL) task \cite{sultani2018real}. Each video is divided into a number of fragments such that  each fragment consists of several consecutive frames.
A bag of fragments is created using a complete set of fragments from a single video. Training of the network is then carried out by defining a ranking loss between two top-scoring fragments, one from the anomalous and the other from the normal bag. However, this approach necessitates to convert each video of the dataset into the same number of fragments which may not always be an appropriate choice. 

Since the real-world datasets contain significantly varying length of videos, fixed number of fragments may not be able to represent events happening over a short span of time in lengthy videos. This problem has been addressed by Zhong \etal \cite{zhong2019graph} who proposed an anomaly detection approach using weakly labeled videos. In their approach, training is performed using noisy labels, where the noise refers to normal fragments within anomalous videos. 
They take advantage of an action classification model \cite{carreira2017quo} to train a graph convolution network \cite{zhong2019graph}, which then helps in cleaning noisy labels from the anomalous videos. The essence of our proposed approach may be considered similar to theirs because we also attempt to mitigate noisy labels. However, our formulation of the problem is entirely different. In our architecture, instead of relying on action classifiers or graph convolution network, we propose to employ a binary clustering based self-reasoning approach which not only attempts to remove noisy labels but also contributes in enhancing the performance of our base framework through a clustering distance loss. 
The main contributions of the current work are summarized below:
\begin{itemize}
  \item Our framework trains in a weakly supervised manner using only video-level annotations to localize anomalous events.
 \item We propose a clustering based self-reasoning approach to clean noise from the labels of anomalous videos. Our approach encourages the FC network to improve clusters, which in return enhances the capability of the network to discriminate anomalous portions of a video. Thus, enabling both the network and the clustering  to complement each other during training.
\item Our method outperforms existing SOTA by yielding frame-level AUC of 79.54\% on UCF-crime, 84.16\% on ShanghaiTech, and 94.47\% on UCSD Ped2 datasets.
\end{itemize}

\section{Proposed Architecture}
The overall proposed framework is visualized in Fig. \ref{fig:architecture}, and all of its components are discussed below: \\
\noindent\textbf{Group of Video Fragments:}
All frames from a video $V_i$ are divided into a group of fragments in such a way that each fragment $F_{(i,j)}$ contains $k$ non-overlapping frames, where $i\in[1,n]$ is the video index in the dataset of $n$ videos and $j\in[1,m_i]$ is the index of $m_i$ fragments in $V_i$. For each video, only one binary label \{normal = 0, anomalous = 1\} is provided.\\
\noindent\textbf{Feature Extractor:}
Spatio-temporal features of each fragment $F_{(i,j)}$ are computed by employing a pre-trained feature extractor model such as Convolution 3D (C3D) introduced by Tran et al. \cite{tran2015c3d}. Our proposed framework is generic and may employ any spatio-temporal feature extractor.\\
\noindent\textbf{Fully Connected Base Network:}
Our base network consists of two fully connected (FC) layers, each followed by a ReLU activation function and a dropout layer, as shown in Fig. \ref{fig:architecture}(d). The input layer receives a spatio-temporal feature vector and the output layer produces an anomaly regression score in the range of $[0, 1]$ through a Sigmoid activation function, where 0 represents a normal fragment and 1 represents an anomalous fragment.\\
\noindent\textbf{Clustering Based Self-Reasoning:}
Given that anomaly detection is a binary problem, clustering algorithms such as k-means \cite{kanungo2002kmeans} can be employed to distribute all fragments into two clusters. These clusters are created using the feature representations of each fragment taken from the output of FC-1 layer.
Clustering here serves two purposes: 1) It helps in generating fragment-level pseudo annotations from video-level labels. 2) It encourages the network to push both clusters away in the case of an anomalous video, and brings both clusters closer in the case of a normal video.

\vspace{-2mm}
\subsection{Training}
As explained previously, our architecture is trained using only video-level labels, therefore we propose a self-reasoning-enabled approach to create pseudo annotations. Moreover, our configuration also encourages FC network and clustering to complement each other towards improving the results throughout training iterations.

\noindent\textbf{Creating Pseudo Annotations:}
For the normal labeled videos, as no anomaly is present, each fragment of these videos can simply be annotated as normal. However, in the case of anomalous videos, several normal fragments may also be present. To handle this, pseudo annotations $y^p_{(i,j)}$ are generated for the anomalous videos. Overall, for a given video $V_i$, the fragment-level labels $ y_{(i,j)} \in \mathbf{y}_{(i)} \in \{0, 1\}^{m_i}$ for training are given as: 
\begin{equation}
    y_{(i,j)}=
    \begin{cases}
    0, & \text{if $V_i$ is normal}\\
    y^p_{(i,j)}, & \text{if $V_i$ is anomalous.} 
    \end{cases}
    \label{eq:labels}
\end{equation}
To obtain $y^p_{(i,j)}$, all fragments from an anomalous video are divided into two clusters assuming one cluster would contain normal while the other would contain anomalous fragments. At this point, one of the two cases may occur: Case 1) cluster A contains most of the normal fragments while cluster B contains most of the anomalous fragments. Case 2) cluster B contains most of the normal fragments while cluster A contains most of the anomalous fragments. It is important to determine the exact case so that the appropriate pseudo-labels may be assigned. In order to do so, a similarity score $s_1$ is computed between the labels predicted by the FC network, $\hat{\mathbf{y}}_{(i)} \in [0,1]^{m_i}$, and the labels generated by clustering, $\mathbf{y}^c_{(i)} \in \{0,1\}^{m_i}$ for all fragments of $V_i$,
\begin{equation}
s_1 =  \frac{\hat{\mathbf{y}}_{(i)}\cdot \mathbf{y}^c_{(i)}}{ ||\hat{\mathbf{y}}_{(i)}||_2 ||\mathbf{y}^c_{(i)}||_2}.
\label{eq:s_1}
\end{equation}
To resolve the ambiguity, another similarity score $s_2$ is also computed between $\hat{\mathbf{y}}_{(i)}$ and the inverted clusters' labels $\neg \mathbf{y}^c_{(i)}$, where $\neg$ is the logical negation:
\begin{equation}
s_2 = \frac{\hat{\mathbf{y}}_{(i)} \cdot \neg\mathbf{y}^c_{(i)}}{||\hat{\mathbf{y}}_{(i)}||_2||\neg\mathbf{y}^c_{(i)}||_2}.
\label{eq:s_2}
\end{equation}
Finally, given $j^{th}$ fragment $F_{(i,j)}$ in $V_i$, the pseudo-label is given as:
\begin{equation}
    y^p_{(i,j)}=
    \begin{cases}
    y^c_{(i,j)}, & \text{if } s_1 \geq s_2\\
    \neg y^c_{(i,j)}, & \text{otherwise}, 
    \end{cases}
    \label{eq:yp}
\end{equation}
Thus, we force the FC network to learn normal fragments in normal videos which are noise-free. For the case of anomalous videos, we find a cluster label configuration which is in accordance with the labels predicted by the FC network. Though, the final fragment labels in an abnormal video are decided by the cluster labels. The pseudo labels thus found are then used for the computation of network loss function.


\subsection{Training Losses}
Overall, our network is trained to minimize the loss: 
\begin{equation}
    L= L_r + \lambda L_c,
    \label{eq:lambda_equation}
\end{equation}
where $L_r$ is given as:
\begin{equation}
L_r = \frac{1}{m_i}\sum_{j=1}^{m_i} ((y_{(i,j)} - \hat{y}_{(i,j)}))^2 
\end{equation}
and $\lambda$ is a trade-off hyperparameter.
Moreover, clustering distance loss, $L_c$, is defined as:
\begin{equation}
    L_c=
    \begin{cases}
    \text{min}(\alpha, d_i), & \text{if $V_i$ is normal}\\
    \frac{1}{d_i}, & \text{if $V_i$ is anomalous,} 
    \end{cases}
    \label{eq:alpha_equation}
\end{equation}
\revision{where $d_i = ||C_1 - C_2||_2$ is the distance between the cluster centers $C_1$ and $C_2$  obtained using k-means algorithm over the fragments of $V_i$, and $\alpha$ is an upper bound on the distance loss}.

\begin{table}[t]
\caption{Frame-level AUC \% performance comparison of our approach with SOTA methods on UCF-crime, ShanghaiTech, and UCSD Ped2 Datasets. The best and the second best performances in each column are shown in red and blue colors.}
\begin{center}
\begin{tabular}{c|c|c|c} 
\text{\textbf{Method}} & \text{\textbf{UCF-crime}} & \text{\textbf{ShanghaiTech}} & \textbf{UCSD Ped2}\\ \hline
Binary SVM \cite{sultani2018real}   & 50.0 & - & - \\ \hline
Hasan \etal \cite{hasan2016anomaly}   & 50.60 & - & - \\ \hline
Lu  \etal \cite{lu2013abnormal}     & 65.51 & - & -  \\ \hline
Sultani  \etal \cite{sultani2018real}   & 75.41 &- & -\\ \hline
TCN-IBL \cite{zhang2019temporal}   & 78.66 & - & - \\ \hline
Adam \etal \cite{adam2008robust} &-&- & 63.0  \\ \hline
MDT \cite{mahadevan2010anomaly} &-&- & 85.0  \\ \hline
SRC \cite{cong2011sparse} &-&- & 86.1  \\ \hline
AL \cite{he2018anomaly} &-&- & 90.1  \\ \hline
AMDN \cite{xu2017detecting} &-&- & 90.8  \\ \hline
Zhong  \etal \cite{zhong2019graph} &  \bestperformance{81.08} & \secondbestperformance{76.44} &\secondbestperformance{93.20}\\ \hline\hline
\textbf{Ours} & \secondbestperformance{79.54} & \bestperformance{84.16} & \bestperformance{94.47} 
\end{tabular}
\end{center}
\label{tab:UCF_crime_AUC}
\vspace{-5mm}
\end{table}

\section{Experiments}
\subsection{Datasets}
Proposed framework is evaluated on three publicly available anomalous events detection datasets including UCF-crime, ShanghaiTech and UCSD Ped2. \textbf{UCF-crime} \cite{sultani2018real} is a weakly labeled anomalous event detection dataset obtained from real-world surveillance videos. For training, it contains 810 videos of anomalous and 800  of normal classes. For testing, it contains 140 anomalous and 150 normal videos.
\textbf{ShanghaiTech} \cite{luo2017shanghaitech} is also an anomalous event dataset recorded in a university campus. We follow a recent split introduced by Zhong \etal\cite{zhong2019graph} containing 63 anomalous and 175 normal videos for training and 44 anomalous and 155 normal videos for testing.  
\textbf{UCSD Ped2} \cite{chan2008ucsd} dataset comprises of 16 normal and 12 anomolous videos. Pedestrians dominate most of the frames whereas anomalies include skateboards, vehicles, bicycles, etc. Following the protocol proposed by He \etal \cite{he2018anomaly}, we randomly selected 10 videos for training (4 normal and 6 anomolous) and 18 videos for testing (12 normal and 6 anomolous). To remove the bias in random selection, experiment is repeated five times and the average performance is reported.

Following existing SOTA \cite{zhong2019graph,sultani2018real}, Area Under the Curve (AUC) of the  ROC curve for frame-level performance is used as an evaluation metric. For the training of backbone network, Adam optimizer \cite{kingma2014adam} is used with a learning rate of 5 $\times$ $10^{-5}$. In Eqs. \eqref{eq:lambda_equation} and \eqref{eq:alpha_equation}, $\lambda$ and $\alpha$ are set to 0.05  and 1 respectively. Our proposed framework is generic and may use several types of feature extractors, however in accordance with the existing techniques, we employ C3D architecture \cite{tran2015c3d}. In C3D, the number of frames per fragment, $k$, is set to 16.

\vspace{-2mm}
\subsection{Comparisons with Existing State-of-the-Art Approaches}
Table \ref{tab:UCF_crime_AUC} summarizes a comparison of our approach with the existing state-of-the-art methods on all three datasets. For the case of UCF-crime, our approach demonstrates superior performance compared to the methods including Sultani \etal \cite{sultani2018real}, Hasan \etal \cite{hasan2016anomaly}, Lu \etal \cite{lu2013abnormal}, Zhang \etal \cite{zhang2019temporal} with a significant margin while comparable results to Zhong \etal \cite{zhong2019graph}. For ShanghaiTech dataset, using the same splits and C3D features, our framework outperforms Zhong \etal \cite{zhong2019graph} by a significant margin of 7.72\% in AUC. For UCSD Ped2 dataset, again our proposed model outperforms most compared methods by a significant margin.
On the average taken over all three reported datasets, our proposed framework obtains 86.06\% AUC, while Zhong \etal \cite{zhong2019graph} obtains an average performance of 83.57\% which demonstrates the overall superiority of our proposed approach. 

\begin{table}[t]
\caption{Top-down ablation study of our proposed approach to observe the importance of different components in terms of AUC \%.}
\begin{center}
\begin{tabular}{c|c|c|c}
                           & \multicolumn{2}{c}{\textbf{Datasets}} \\ \hline
\textbf{Method}                  & \textbf{ShanghaiTech}   & \textbf{UCF-crime} & \textbf{UCSD Ped2}   \\ \hline
FC + $L_c$ + $y^p$              & \bestperformance{84.16}           & \bestperformance{79.54} & \bestperformance{94.47}  \\ \hline
FC + $y^p$  & \secondbestperformance{83.37}            & \secondbestperformance{77.13}  & \secondbestperformance{91.34}       \\ \hline
FC + $L_c$         & 81.65            & 76.59   &   89.1   \\ 

\end{tabular}
\end{center}
\label{tab:ablation}
\vspace{-5mm}
\end{table}

\begin{figure*}[t]
\begin{center}
\includegraphics[width=.85\linewidth]{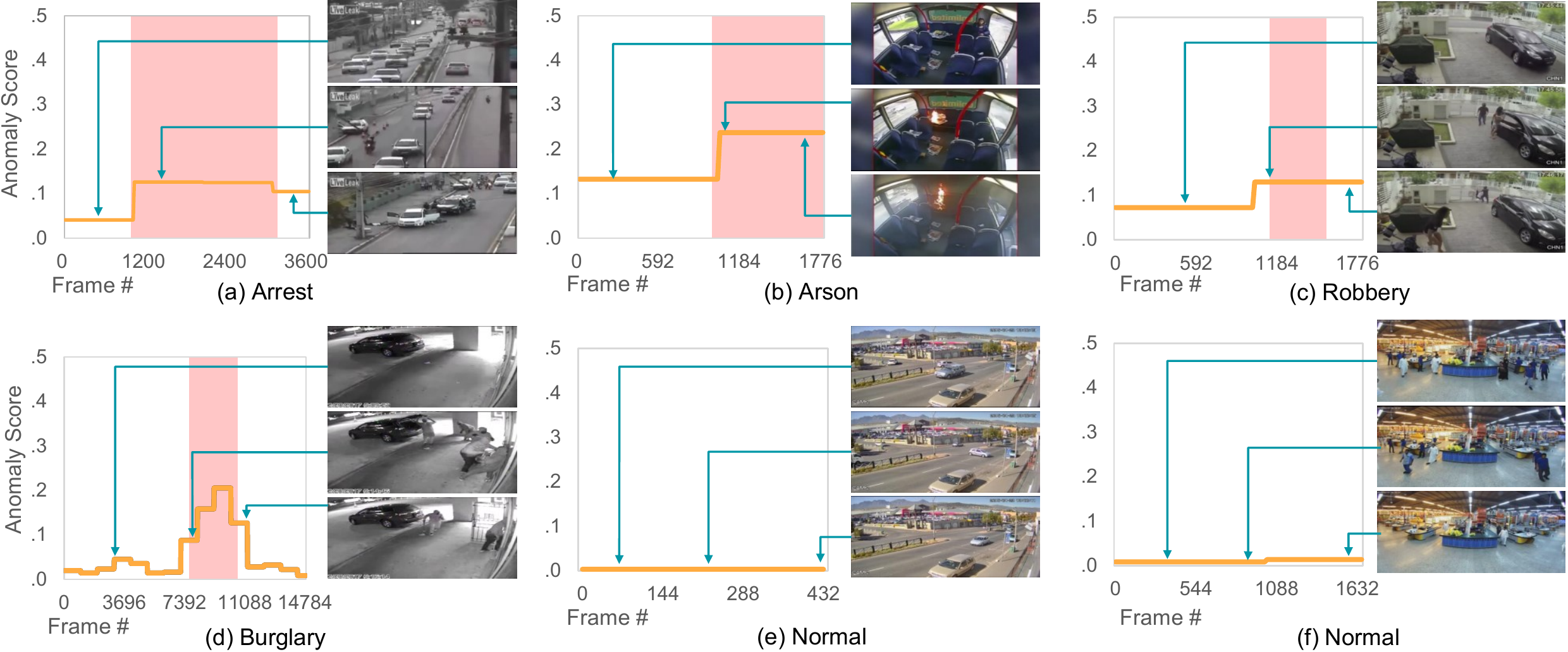}
\end{center}
\vspace{-3mm}
  \caption{Qualitative results of our approach on test videos of UCF-crime dataset. Colored rectangular window represents anomaly ground truth. (a),(b),(c), and (d) show the scores predicted by our model on anomalous whereas (e) and (f) show the scores on normal videos.}
\label{fig:qualitative}
\end{figure*}

\begin{figure*}[t]
\begin{center}
\includegraphics[width=.90\linewidth]{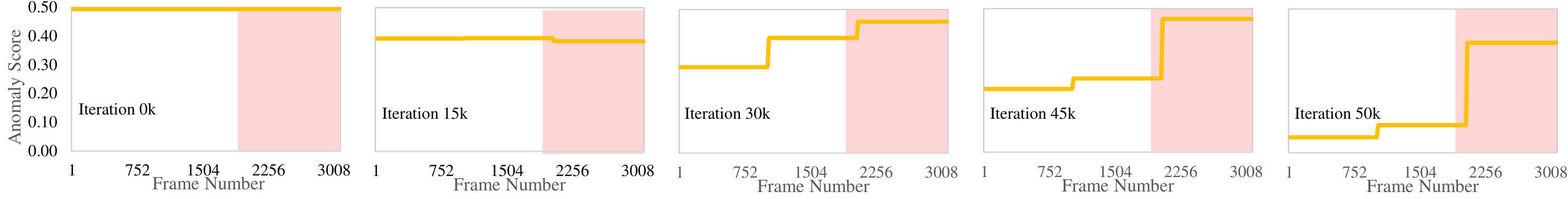}
\end{center}
\vspace{-3mm}
  \caption{Visualization of the frame-level anomaly scores produced by our model over after several training iterations on a test video taken from fight class of the UCF-crime dataset. Due to the self-reasoning architecture, our network evolves to produce higher scores in the anomalous regions of the video.}
\label{fig:scores_evolution}
\end{figure*}

\begin{figure}[bh!]
\begin{center}
\includegraphics[width=.65\linewidth]{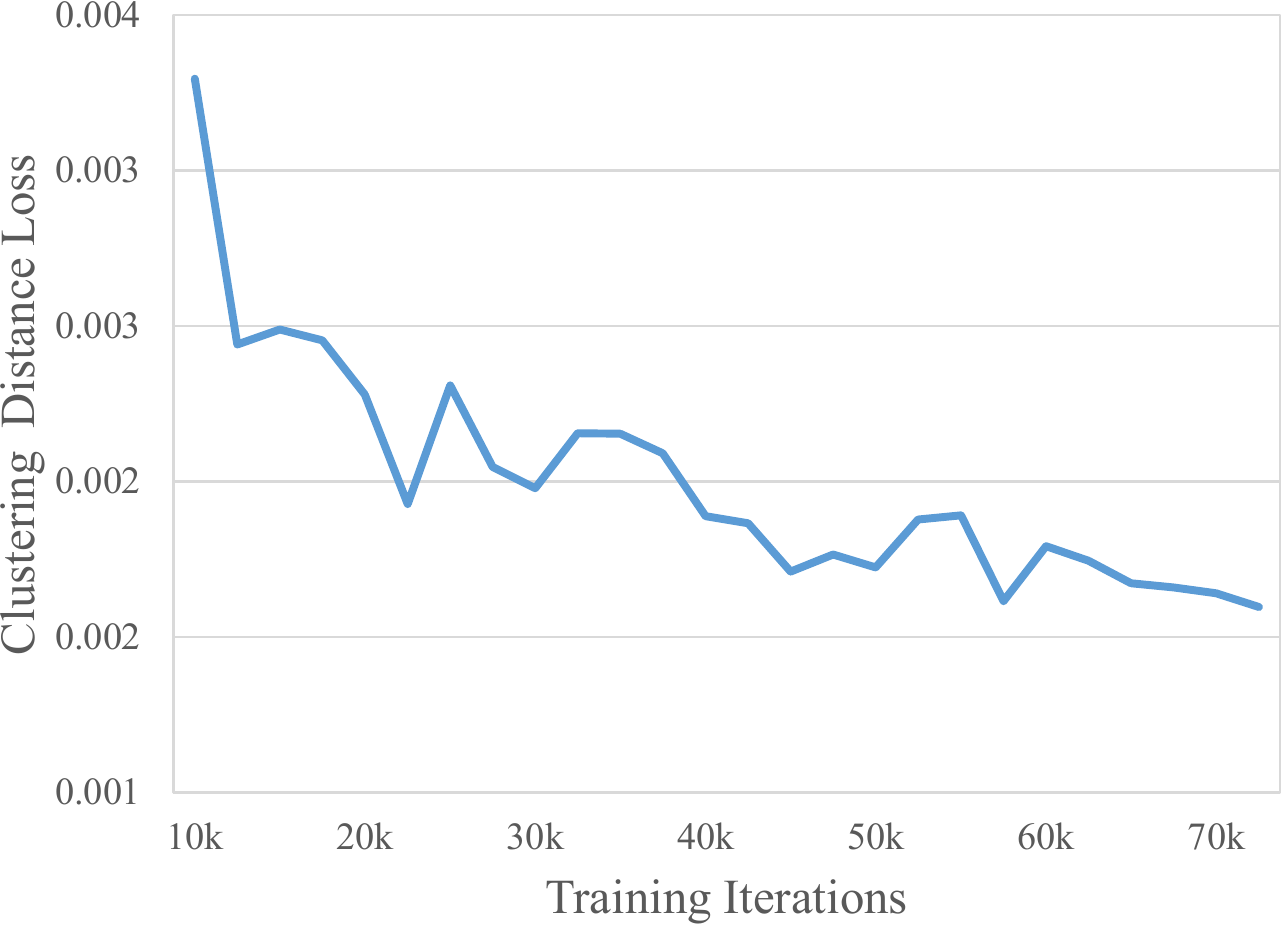}
\end{center}
\vspace{-3mm}
  \caption{Clustering distance loss over several training iterations plotted on the UCF-crime dataset.}
\label{fig:clusters_loss_plot}
\end{figure}

\vspace{-3mm}
\subsection{Ablation Study}
A detailed ablation study on all three datasets including UCF-crime, ShanghaiTech and UCSD Ped2,  is provided in Table \ref{tab:ablation}. We follow a top-down approach in which different components of the proposed framework are systematically removed to observe the loss of performance compared to the full system. In the case of ShanghaiTech dataset, removal of clustering distance loss ($L_c$) resulted in a drop of 0.79\% whereas removal of clustering based pseudo annotations, $\mathbf{y}^p$ for anomalous videos, resulted in a drop of  2.51\%. Note that when we remove $\mathbf{y}^p$, the labels of all fragments of an anomalous video are set to 1 in Eq. \eqref{eq:labels}.
Experiments on UCF-crime and UCSD Ped2 datasets also demonstrated similar trends (Table \ref{tab:ablation}).

\vspace{-2mm}
\subsection{Qualitative Results}
Anomaly score plots of several normal and anomalous test videos from the UCF-crime dataset are visualized in Fig. \ref{fig:qualitative}. Overall, our network produces distinctive scores for anomalous portions of the videos. For the case of Arrest event (Fig. \ref{fig:qualitative} (a)), a significant increase in the anomaly score shows the start of the anomalous event. At the end of this event, the anomaly score still retains higher value. It is because even after the arrest has been made, the cars parked on roadside still represent some extent of abnormality. A similar behavior can also be observed in the robbery event (Fig. \ref{fig:qualitative} (c)) where the annotated part of the event is smaller than the actual event. For the case of burglary (Fig. \ref{fig:qualitative} (d)), our framework detects quite accurately the start and the end of the anomalous event. For the two normal cases (Figs. \ref{fig:qualitative} (e) \& (f)), our proposed framework outputs significantly low scores, making it easy to discriminate anomalous events from the normal events.

Due to the self-reasoning, our proposed system evolves over iterations on the training videos to produce higher anomaly scores in the abnormal regions despite using only video-level labels. Fig. \ref{fig:scores_evolution} shows such evolution of the system on a test video from UCF-crime dataset. As the number of iterations increases, the difference between the anomaly score over normal and anomalous regions also increases. 
Fig. \ref{fig:clusters_loss_plot} shows the reduction in clustering distances loss over consecutive training iterations. It demonstrates that the backbone network is learning to produce better internal representations, causing reduction in inter-cluster distance for the case of normal and increase in this distance for the case of anomalous videos. 

\vspace{-2mm}
\section{Conclusion}
A weakly-supervised approach is proposed to learn anomalous events using video-level labels.
 Compared to using frame-level annotations, the video-level annotations contain significant noise in case of anomalous videos.
To this end, binary clustering is employed which enables self-reasoning to mitigate these noisy labels. The proposed framework enables both the FC network and the clustering algorithm to complement each other in improving the quality of results. Our method demonstrates state-of-the-art results by yielding 79.54\%,  84.16\% and 94.47\% frame-level AUC on  UCF-crime, ShanghaiTech and UCSD Ped2 datasets.

\balance
\bibliographystyle{IEEEtran} 
\bibliography{citations}
 
\end{document}